\pdfoutput=1

\documentclass[11pt]{article}

\usepackage{acl}

\usepackage{times}
\usepackage{latexsym}

\usepackage[T1]{fontenc}

\usepackage[utf8]{inputenc}

\usepackage{microtype}
\usepackage{multirow}
\usepackage{booktabs} 
\usepackage{graphicx}
\usepackage[export]{adjustbox}

\usepackage{amsmath}

\title{Query2Particles: Knowledge Graph Reasoning with Particle Embeddings}

\author{Jiaxin Bai$^1$, Zihao Wang$^1$, Hongming Zhang$^2$, and Yangqiu Song$^1$ \\
  $^1$CSE, HKUST \\
  $^2$Tencent AI Lab \\
  \texttt{\{jbai, zwanggc, yqsong\}@cse.ust.hk, hongmzhang@tencent.com} \\}

\begin{document}
\maketitle
\begin{abstract}
Answering complex logical queries on incomplete knowledge graphs (KGs) with missing edges is a fundamental and important task for knowledge graph reasoning. The query embedding method is proposed to answer these queries by jointly encoding queries and entities to the same embedding space. Then the answer entities are selected according to the similarities between the entity embeddings and the query embedding. As the answers to a complex query are obtained from a combination of logical operations over sub-queries, the embeddings of the answer entities may not always follow a uni-modal distribution in the embedding space. Thus, it is challenging to simultaneously retrieve a set of diverse answers from the embedding space using a single and concentrated query representation such as a vector or a hyper-rectangle. To better cope with queries with diversified answers, we propose Query2Particles (Q2P), a complex KG query answering method. Q2P encodes each query into multiple vectors, named particle embeddings. By doing so, the candidate answers can be retrieved from different areas over the embedding space using the maximal similarities between the entity embeddings and any of the particle embeddings. Meanwhile, the corresponding neural logic operations are defined to support its reasoning over arbitrary first-order logic queries. The experiments show that Query2Particles achieves state-of-the-art performance on the complex query answering tasks on FB15k, FB15K-237, and NELL knowledge graphs. 
\end{abstract}

\section{Introduction}

\begin{figure}[t]
\begin{center}
\includegraphics[clip, trim=6cm 6.3cm 6.8cm 4.2cm,width=\linewidth]{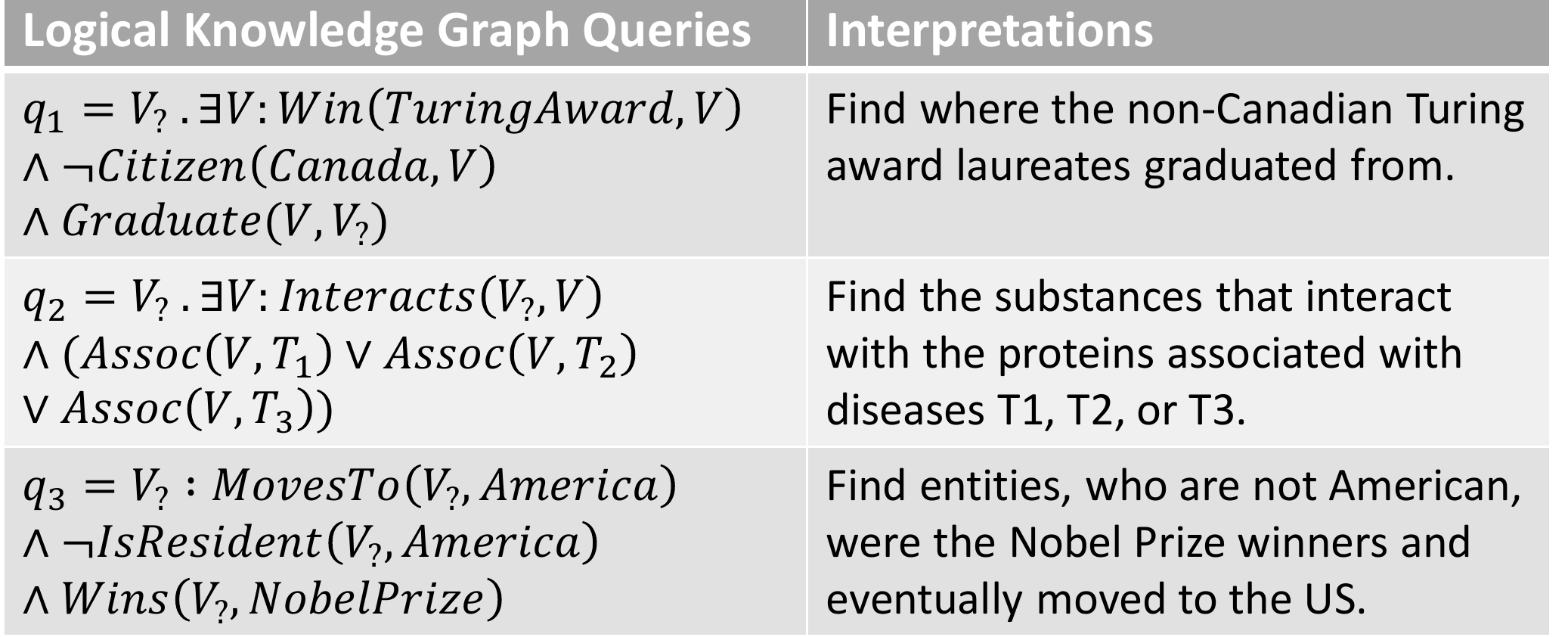}
\end{center}
\vspace{-0.3cm}
\caption{
The example logical knowledge graph queries and their interpretations in natural language. 
}
\vspace{-0.4cm}
\label{fig:query_examples}
\end{figure}

Reasoning over a factual knowledge graph  (KG) is the process of deriving new knowledge or conclusions from the existing data in the knowledge graph \cite{chen2020review}.
A recently developed sub-task of knowledge graph reasoning is complex query answering, which aims to answer complex queries over large knowledge graphs \citep{hamilton2018embedding,ren2020query2box, ren2020beta}. 
Compared to KG completion tasks \cite{zhiyuan2016knowledge, west2014knowledge}, complex query answering requires reasoning over multi-hop relations and logical operations.
As shown in Figure \ref{fig:query_examples}, complex KG queries are
defined
in predicate logic forms with relation projection operations, existential quantifiers $\exists$, logical conjunctions $\land$, disjunctions $\lor$, and negation $\lnot$. 
Answering these queries is challenging because real-world knowledge graphs (KG), such as Freebase \citep{bollacker2008freebase}, NELL \citep{carlson2010toward}, and DBPedia \citep{bizer2009dbpedia}, are incomplete.
Consequently, sub-graph matching methods cannot be used to find the answers. 

\begin{figure*}[t]
\begin{center}
\includegraphics[clip, trim=2cm 11.3cm 1cm 1.5cm,width=\linewidth]{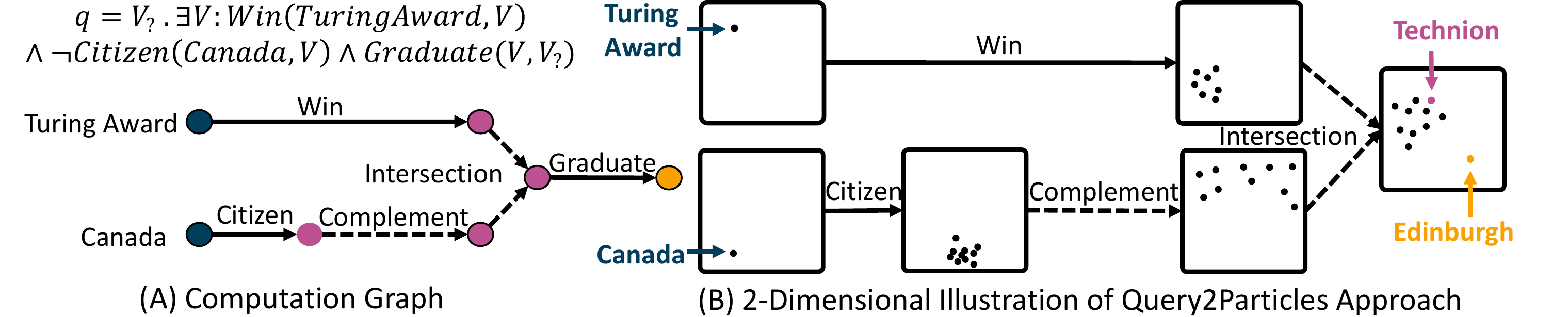}
\end{center}
\vspace{-0.3cm}
\caption{
An example of answering a knowledge graph query by using the Query2Particles method. (A) The computational graph corresponds to the query ``where did the non-Canadian Turing award laureates graduate from.''
(B) The Query2Particles encodes each query into a set of vectors, called particle embeddings. 
The logical operations iteratively compute particle embeddings following the computational graph. 
The answers are determined by using the maximum similarities between the entity embeddings and any one of the resulting particle embeddings. 
}
\vspace{-0.3cm}
\label{fig:computation_graph}
\end{figure*}

To address the challenge raised from the incompleteness of knowledge graphs, the query embedding methods are proposed \citep{hamilton2018embedding,ren2020query2box, ren2020beta,  sun2020faithful}.   
In this line of research, the queries and entities are jointly encoded into the same embedding space, and the answers are retrieved based on similarities between the query embedding and entity embeddings. 
In general, there are two steps in encoding a query to the vector space. 
First, a query is parsed into a computational graph with a directed acyclic graph (DAG) structure, as shown in Figure \ref{fig:computation_graph} (A).
Then, the query representation is iteratively computed following the neural logic operations and relation projections in the DAG.  


Although the query embedding methods are robust for dealing with the incompleteness of KGs, the embedding structure used for encoding the queries can be improved.
Because of the multi-hop and compositional nature of complex KG queries, a single query may contain multiple sufficiently diverse answers. 
Thus, the ideal query embedding may follow a multi-modal distribution\footnote{A multi-modal distribution is a distribution with two or more distinct peaks in the probability density function.} in the embedding space.
For example, the answers to the query, ``{\it Find entities, who are not American, were the Nobel Prize winners and eventually moved to the US},'' involve intermediate entities with different attributes, such as gender, nationality, research fields, etc. 
It is difficult to use a single embedding vector to find all final answer embeddings.
Box embedding~\cite{ren2020query2box} partially solved this problem, but for complicated attributes, a single box may be too coarse, and intermediate entities are distributed far away from each other, 
so they are more like several disjoint clusters rather than a single big region in the embedding space.
So for the query embedding methods, the capability to simultaneously encode 
a set of answers from different areas is necessary.

To better address the diversity of answers, 
we propose Query2Particles, a new query embedding method for complex query answering. 
In this approach, each query is encoded into a set of vectors in the embedding space, called particle embeddings. 
The particle embeddings of a query are iteratively computed by following the computational graph parsed from the query. 
Then the answers to this query are determined by using the maximum similarities between the entity embeddings and any one of the resulting particle embeddings. 
Experimental results show that Query2Particle achieves state-of-the-art performance on complex query answering over three standard knowledge graphs: FB15K, FB15k-237, and NELL.
Meanwhile, the inference speed of Query2Particles is comparable to other query embedding methods and is higher than query decomposition methods on multi-hop queries. 
Further analysis indicates that the optimal numbers of particles for different query types depend on the structures of the queries. 
Our experimental code is released on github\footnote{https://github.com/HKUST-KnowComp/query2particles}. 

\begin{figure}[t]
\begin{center}
\includegraphics[clip, trim=0.5cm 9.5cm 0cm 1.5cm,width=\linewidth]{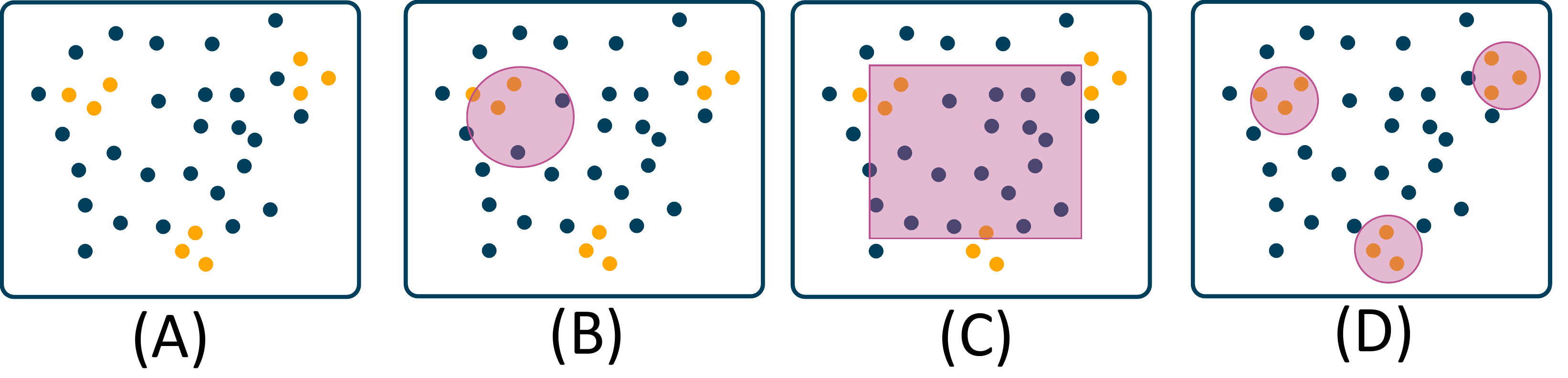}
\end{center}
\vspace{-0.2cm}
\caption{
In the example embedding space, the yellow dots are the answer entities, and the blue dots are the non-answer entities.
The purple areas in (B), (C), and (D) demonstrate the neighborhoods of the vector embedding, the box embedding, and the particle embeddings respectively. In this case, the particle embeddings are more suitable for finding the answers clustered in different areas in the embedding space. 
}
\vspace{-0.2cm}
\label{fig:particle_illustration}
\end{figure}

\section{Related Work}

Other query embedding approaches are closely related to our work. 
These query embedding methods leverage different structures to encode logical KG queries, and they can answer various scopes of logical queries. 
The GQE method proposed by \citet{hamilton2018embedding} can answer the conjunctive queries by representing queries as vector representations. 
\citet{ren2020query2box} used hyper-rectangles to encode and answer existential positive first-order (EPFO) queries. 
At the same time, \citet{sun2020faithful} proposed to improve the faithfulness of the query embedding method by using centroid-sketch representations on EPFO queries. 
The conjunctive queries and EPFO queries are both subsets of first-order logic (FOL) queries. 
The Beta Embedding \citep{ren2020beta} is the first query embedding method that supports a full set of operations in FOL by encoding entities and queries into probabilistic Beta distributions.
In a contemporaneous work, \citet{zhang2021cone} uses cone embeddings to encode the FOL queries.
As shown in Figure \ref{fig:particle_illustration}, compared to these query embedding approaches, the Q2P method can encode the FOL queries to address the diversity of answers. 
Note that, \citet{ren2020query2box} proposed to use the disjunctive normal form (DNF) to address the answer diversities resulting from the union operations.
This partly solve the problem, but
the diversity of the answers is not solely caused by the union operation, but a joint effort of multi-hop projections, intersection, and complement.
As a result, using particle embeddings is a more general solution.

Query decomposition \cite{arakelyan2020complex} is another approach to answering complex knowledge graph queries.
In this line of research, a complex query is decomposed into atomic queries, and the probabilities of atomic queries are modeled by link predictors. In the inference process, continuous optimization and beam search are used for finding the answers. 
Meanwhile, the rule and path-based methods \cite{guo2016jointly,xiong2017deeppath,lin2018multi,guo2018knowledge,chen2019embedding} use pre-defined or learned rules to do multi-hop KG reasoning. 
These methods explicitly model the intermediate entities in the query.
Instead, the query embedding methods directly embed the complex query and retrieve the answers without explicit modeling intermediate entities. 
So the query embedding methods are more scalable to large knowledge graphs and complex query structures. 

Neural link predictors \cite{wang2014knowledge, trouillon2016complex,dettmers2018convolutional, sun2018rotate} are also related to this work. 
The link predictors learn the distributed representations of entities and relations in embedding space and use different neural structures to classify whether there exists a certain relation between two entities. 
The link predictors can be used for one-hop queries, but cannot be directly used for answering complex queries. 

\section{Preliminaries}

In this section, we formally define the complex logical knowledge graph queries and the corresponding computational graphs.
The knowledge graph reasoning is conducted on a multi-relational knowledge graph $\mathcal{G} = (\mathcal{V}, \mathcal{R})$, where each vertex $v \in \mathcal{V}$ represents an entity, and each relation $r \in \mathcal{R} $ is a binary function defined as $r: \mathcal{V}\times \mathcal{V} \rightarrow \{0, 1\}$. 
For any $r\in \mathcal{R}$, and $u,v\in \mathcal{V}$, there is a relation $r$ between entities $u$ and $v$ if and only if $r(u, v) = 1$.

\subsection{First-Order Logic Query}
The complex knowledge graph query is defined in first-order logic form with logical operators such as existential quantifiers $\exists$, conjunctions $\land$, disjunctions $\lor$, and negations $\lnot$. 
In a first-order logic query, there is a set of anchor entities $V_a \in \mathcal{V}$, existential quantified variables $V_1, V_2, ... V_k \in \mathcal{V}$, and a unique target variable $V_?\in \mathcal{V}$. 
The query intends to find the answers $V_?\in \mathcal{V}$, such that there simultaneously exist $V_1, V_2, ... V_k \in \mathcal{V}$ satisfying the logical expression in the query. 
For each FOL query, it can be converted to a disjunctive normal form, where the query is expressed as a disjunction of several conjunctive expressions:
\begin{align}
q[V_?] &= V_? . \exists V_1, ..., V_k: c_1 \lor c_2 \lor ... \lor c_n ,\label{equa:definition} \\ 
c_i &= e_{i1} \land e_{i2} \land ... \land e_{im} .
\end{align}
Each $c_i $ represents a conjunctive expression of several literals $e_{ij}$, and each
$e_{ij}$ is an atomic or the negation of an atomic expression expressed by any of the following expressions:
$e_{ij} = r(v_a, V)$, 
$e_{ij} = \lnot r(v_a, V)$, 
$e_{ij} = r(V, V')$, or 
$e_{ij} = \lnot r(V, V')$.
Here $v_a \in V_a$ is one of the anchor entities, and $V,V' \in \{ V_1, V_2, ... ,V_k, V_? \}$ are distinct variables satisfying $V \neq V'$. 

\subsection{Computational Graph and Operations}
As shown in Figure \ref{fig:computation_graph} (A), for a first-order query, there is a corresponding computational graph. 
In the computational graph, each node corresponds to an intermediate query embedding, and each edge corresponds to a neural logic operation to be defined in the following section. 
Both the input and output of these operations are query embeddings. 
These operations are used for implicitly modeling different set operations over the intermediate answer sets. 
These set operations include relational projection, intersection, union, and complement: (1) \textit{Relational Projection}: 
Given a set of entities $A$ and a relation $r \in R$, the relational projection will return all entities having relation $r$ with at least one of entity $e \in A$. Namely, $P_r(A) = \{ v \in \mathcal{V} | \exists v' \in A, r(v', v) = 1 \}$;
(2) \textit{Intersection}: 
Given sets of entities $A_1,...A_n \subset \mathcal{V}$, this operation computes their intersection $ \cap_{i=1}^{n} A_i $;
(3) \textit{Union}: 
Given several sets of entities $A_1,...A_n \subset \mathcal{V}$, the union operation calculates their union $\cup_{i=1}^{n} A_i$;
(4) \textit{Complement}:
Given a set of entities $A$, the complement operation calculates its absolute complement $\mathcal{V} - A$.

\begin{figure*}[t]
\begin{center}
\includegraphics[clip, trim=0.5cm 9.8cm 1.3cm 2.1cm,width=\linewidth]{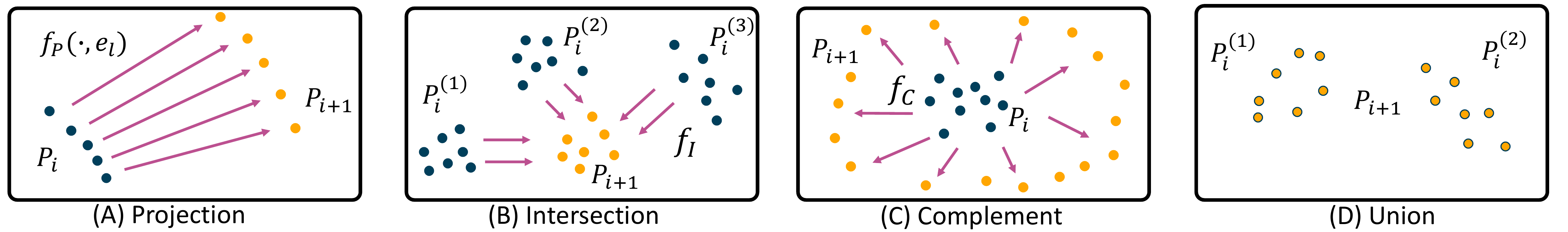}
\end{center}
\vspace{-0.3cm}
\caption{ The $P_i$ in blue and the $P_{i+1}$ in yellow are the input and output particles respectively. 
(A) The embeddings $P_i$ are projected to $P_{i+1}$ by the relation  $l$. 
(B) The resulting embeddings $P_{i+1}$ are computed from three sets of particles $P_i$ by the intersection $f_I$. 
(C) The output $P_{i+1}$ are computed from the input $P_{i}$ by the complement $f_C$. 
(D) The $P_{i+1}$ are directly taken from the all input particles in $P_{i}$ without any additional parameterization. 
}
\label{fig:projection}
\vspace{-0.4cm}
\end{figure*}

\section{Query2Particles}
In this section, we first introduce the particle embeddings structure and the neural logic operations, and 
then we present the learning of the model. 

\subsection{Particles Representations of Queries}
In Query2Particles, each query is represented as a set of vectors, called particles. 
For simplicity, a set of particles $\{p^{(k)}\}_{k=1}^{K}$ are represented as a matrix $P$. 
All the operations discussed in the following sections are invariant to the permutations of the particle vectors in the matrix.
Formally, the particle embeddings $P \in R^{d\times K} $ are
\begin{align}
    P = [p^{(1)}, p^{(2)},...,p^{(K)}], \label{equation:particle_definition}
\end{align}
where each vector $p^{(k)} \in R^d $ is a particle vector.
As shown in Figure \ref{fig:computation_graph}, the computations along the computation graph start with the anchor entities, such as ``Turing Award''.  
Suppose the entity embedding of an anchor entity $v$ is denoted as $e_v \in R^d$. 
Then, the initial particle embeddings are computed as
the sum of $e_v$ and a learnable offset matrix $M \in R^{d\times K}$,
\begin{align}
    P_0 = e_v + M.
\end{align}
Here and in the following sections, the addition between the matrix $M$ and the vector $e_v$ is defined as the broadcasted element-wise addition.

\subsection{Logical Operations}

In this sub-section, we define and parameterize four types of neural logic operations: projection, intersection, negation, and union.

\subsubsection{Projection}
Suppose the $e_l \in R^d$ is the embedding vector of the relation $l$. 
The relation projection $f_P$ is expressed as 
$P_{i+1} = f_P(P_i, e_l)$,
where the $P_i$ and $P_{i+1}$ are input and output particle embeddings. 
Instead of directly adding the same relation embedding $e_l$ to all particles in $P_i$ to model the relation projection following \cite{bordes2013translating},  
we incorporate multiple neuralized gates \citep{chung2014empirical} to individually adjust the relation transition for each particle in $P_i$, which are expressed as follows:
\begin{align}
&Z = \sigma(W^{P}_ze_l + U_zP_i  + b_z), \\
&R = \sigma(W^{P}_re_l + U_rP_i + b_r), \\
&T = \phi(W^{P}_he_l + U_h(R \odot P_i ) + b_h),\\
&A_i = (1-Z) \odot P_i + Z \odot T.
\end{align}
Here,  $\sigma$ and $\phi$ are the sigmoid and hyperbolic tangent functions, and  $\odot$ is the Hadamard product.
Also, $W^{P}_z, W^{P}_r, W^{P}_h, U_z, U_r, U_h$ are parameter matrices.
$T$ is interpreted as the relation transitions for each of the particles given the relation embedding $e_l$, and 
$Z$ and $R$ are the update gate and the reset gate used for customizing the relation transitions for each particle. 
Meanwhile, the relation projection result for each particle should also depend on the positions of other input particles. 
To allow information exchange among different particles, a scaled dot-product self-attention \citep{vaswani2017attention} module is also incorporated,
\begin{align}
P_{i+1} = \texttt{Attn}(W^{P}_q A^T_i , W^{P}_k A^T_i, W^{P}_v A^T_i )^T.
\end{align}
The $W^{P}_q, W^{P}_k, W^{P}_v\in R^{d\times d}$ are parameters used for modeling the input Query, Key, and Value for the self-attention module \texttt{Attn}.
The \texttt{Attn} represents the scaled dot-product self-attention,
\begin{align}
    \texttt{Attn}(Q, K, V) = \texttt{Softmax}(\frac{QK^T}{\sqrt{d}}) V.
\end{align}
Here, the $Q$, $K$, and $V$ represent the input Query, Key, and Value for this attention layer.

\subsubsection{Intersection}
The intersection operation $f_I$ is defined on multiple sets of particle embeddings $\{P^{(n)}_i \}_{n=1}^N$. It outputs a single set of particle embeddings $P_{i+1} = f_I(\{P^{(n)}_i \}_{n=1}^N)$.
The particles from the $\{P^{(n)}_i \}_{n=1}^N$ are first merged into a new matrix $P_i = [P^{(1)}_i, P^{(2)}_i, ...,P^{(N)}_i]\in R^{d \times NK}$, 
and this matrix $P_i$ serves as the input of the intersection operation. 
The operation updates the position of each particle according to the positions of other input particles in $\{P^{(n)}_i \}_{n=1}^N$. 
This process is modeled by the scaled dot-product self-attention followed by a multi-layer perceptron (MLP) layer,
\begin{align}
    & A_i = \texttt{Attn}(W^{I}_{q}P^T_i, W^{I}_kP^T_i, W^{I}_vP^T_i )^T, \\
    & P_{i+1}= \texttt{MLP}(A_i).
\end{align}
Here $W^{I}_q,W^{I}_k,W^{I}_v\in R^{d \times d}$ are parameters for the self-attention layer.  
The \texttt{MLP} here denotes a multi-layer proceptron layer with \texttt{ReLU} activation, and the parameters in the \texttt{MLP} layers in different operations are not shared.
To keep the number of particles unchanged, we uniformly sub-sample $K$ particles out of the $NK$ particles in $P_{i+1}$ as the final output of the intersection operation.


\subsubsection{Complement}

The input of the complement operation is a single set of particle embeddings $P_i$, and the operation $f_C$ is formulated as $P_{i+1} = f_C(P_i)$.
The complement operation updates the position of each particle based on the distributions of other input particles. 
The operation is then modeled by scaled dot-product attention followed by an MLP layer, and this can be formulated by

\begin{align}
     & A_i = \texttt{Attn}(W^{C}_{q}P^T_i , W^{C}_kP^T_i, W^{C}_vP^T_i )^T, \\
    & P_{i+1} = \texttt{MLP}(A_i).
\end{align}
Here, the $P_{i+1} \in  R^{d \times K}$ are the resulting particle embeddings for the complement operation, and the values in $W^C_q,W^C_k,W^C_v\in R^{d \times d}$ are parameters.
Intuitively speaking, the proposed structure can model the complement operation by encouraging the particles to move towards the areas that are not occupied by any of the input particles.

\subsubsection{Union}
The union operation is directly modeled by all the input particles without extra parameterization.
In detail, the particles from the input particle embeddings are directly merged into a new set of particles, 
\begin{align}
    f_U(\{P^{(n)}_i\}_{n=1}^N) = [P^{(1)}_i, ...,P^{(N)}_i].
\end{align}

\subsection{Scoring}

After the particle embeddings $P_{T}\in  R^{d \times K}$ for the target variable of the query $q$ are computed, the scoring function $\phi$ between the particle embeddings $P_{T}$ and each entity embedding $e_v$ is used for calculating the maximal similarities between each particle vectors in $\{p^{(k)}_T\}_{k=1}^{K}$ and entity embedding vector. 
Here, the inner product is used to compute the similarity scores between vectors, and the overall scoring function is expressed by 

\begin{align}
\phi(P_T, e_v) = \max_{k\in\{1,2,...,K\}} <p^{(k)}_T,  e_v>.
\end{align}
\subsection{Learning Query2Particles}
To train the Query2Particles model, we compute the normalized probability of the entity $v$ being the correct answer of query $q$ by using the \texttt{softmax} function on all similarity scores,
\begin{align}
    p(v, q) = \frac{\phi(P_T, e_v)}{\sum_{v'\in V} \phi(P_T, e_{v'})  }.
\end{align}
Then we construct the cross-entropy loss from the given probabilities to maximize the log probabilities of all correct query-answer pairs:
\begin{align}
    L = -\frac{1}{N} \sum_i  \log p(v^{(i)}, q^{(i)}).
\end{align}
The $(v^{(i)}, q^{(i)})$ denotes is one of the positive query-answer pairs, and  in total there are $N$ such pairs.

\begin{figure*}[t]
\begin{center}
\includegraphics[clip, trim=1cm 11.8cm 1cm 1.1cm,width=\linewidth]{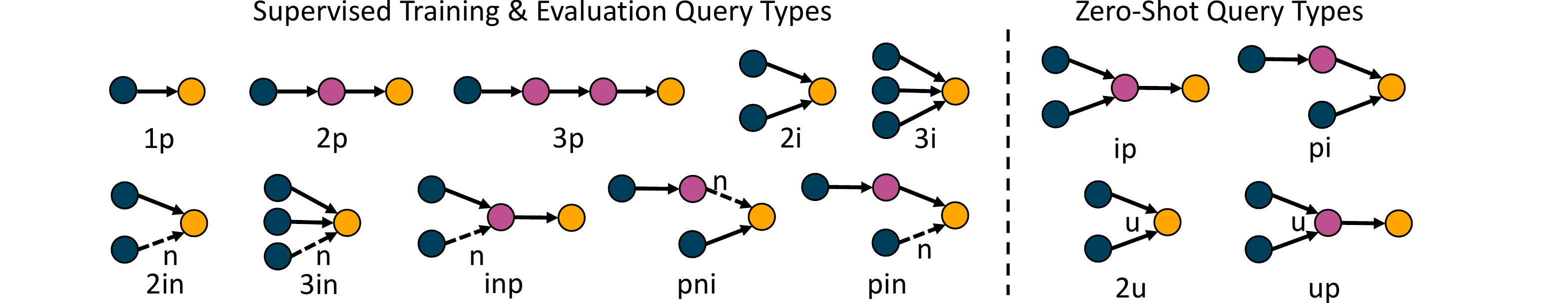}
\vspace{-0.8cm}
\end{center}
\caption{
The query structures used for training and evaluation. For brevity, the $p$, $i$, $n$, and $u$ represent the projection, intersection, negation, and union operations respectively. The query types on the left are trained and evaluated under supervised settings. There are not training queries for the four types of queries on the right, and they are directly evaluated at the test time to measure the generalization capability of the models to unseen query types.
}
\label{fig:query_types}
\end{figure*}

\begin{table*}[t]
\begin{center}
\small
\begin{sc}
\begin{tabular}{p{1.8cm}||p{  0.6cm}p{  0.6cm}p{  0.6cm}|p{  0.6cm}p{  0.6cm}|p{  0.6cm}p{  0.6cm}|p{  0.6cm}p{  0.6cm}|l||c||c}
\toprule
& \multicolumn{10}{c||}{FB15K} &\multicolumn{1}{c||}{FB15K-237}&\multicolumn{1}{c}{NELL}\\
\midrule
Model& 1p&2p&3p&2i&3i&Pi&Ip&2u&Up&Avg&Avg&Avg\\
\midrule
BetaE & 65.1 & {25.7} & {24.7} & {55.8} &  {66.5} &  {43.9} & {28.1} & \textbf{{40.1}} & 25.4 & {41.6} &20.9& 24.6 \\
 Q2B & {68.0} & 21.0 & 14.2 & 55.1 & {66.5} & 39.4 & 26.1 & 35.1  & 16.7 &  38.0 &20.1& 22.9 \\
 GQE  & 54.6 & 15.3 & 10.8 & 39.7 & 51.4 & 27.6 & 19.1 & 22.1  & 11.6  &  28.0 &16.3& 18.6\\
Q2P (Ours)  & \textbf{82.6}  & \textbf{30.8}  & \textbf{25.5}  & \textbf{65.1} & \textbf{74.7}  & \textbf{49.5} & \textbf{34.9}  &32.1& \textbf{26.2}  & \textbf{46.8}  &\textbf{21.9} & \textbf{25.5} \\

\bottomrule
\end{tabular}
\end{sc}
\end{center}
\vspace{-0.3cm}
\caption{The MRR results for existential positive first-order (EPFO) queries used by  \citet{ren2020beta}. The full results are shown in the supplementary materials. 
}
\label{tab:epfo_beta}
 \vspace{-0.3cm}
\end{table*}

\begin{table*}[t]
\begin{center}

\begin{sc}
\small
\begin{tabular}{p{1.8cm}|p{1.8cm}||p{0.40cm}p{0.65cm}|p{0.40cm}p{0.65cm}|p{0.40cm}p{0.65cm}|p{0.40cm}p{0.65cm}|p{0.40cm}p{0.65cm}||p{0.40cm}p{0.65cm}}
\toprule
\multirow{2}{*}{Dataset}&\multirow{2}{*}{Model}& \multicolumn{2}{c}{2in} &\multicolumn{2}{c}{3in}&\multicolumn{2}{c}{Inp}&\multicolumn{2}{c}{Pin}&
\multicolumn{2}{c}{Pni}&\multicolumn{2}{c}{Avg}\\
 & &\multicolumn{1}{p{0.55cm}}{Mrr} & \multicolumn{1}{p{0.55cm}}{H@10} &Mrr & \multicolumn{1}{p{0.55cm}}{H@10} &Mrr & \multicolumn{1}{p{0.55cm}}{H@10} &Mrr & \multicolumn{1}{p{0.55cm}}{H@10} &Mrr & \multicolumn{1}{p{0.55cm}}{H@10} &Mrr & \multicolumn{1}{p{0.55cm}}{H@10} \\
\midrule
\multirow{2}{*}{FB15k}
& BetaE& 14.3 &30.8 & {14.7} &31.9& 11.5 &23.4& 6.5 &14.3& 12.4 &26.3& 11.8 &25.3 \\
& Q2P (Ours) &\textbf{21.9} &\textbf{41.3} &\textbf{20.8} &\textbf{40.2} &\textbf{12.5} &\textbf{24.2} &\textbf{8.9} &\textbf{18.8} &\textbf{17.1}  & \textbf{33.6}  & \textbf{16.4} & \textbf{31.6} \\
\midrule
\multirow{2}{*}{FB15k-237} 
& BetaE  & \textbf{5.1} & \textbf{11.3}& 7.9 & 17.3& 7.4 & 16.0& 3.6 &8.1 & 3.4 &7.0& 5.4 &11.9 \\
& Q2P (Ours) &4.4 &10.1  &\textbf{9.7}  &\textbf{20.7} & \textbf{7.5}&\textbf{16.7}& \textbf{4.6} &\textbf{9.9} & \textbf{3.8} &\textbf{7.2} & \textbf{6.0}  &\textbf{12.9}  \\

\midrule
 \multirow{2}{*}{NELL}
  & BetaE  & \textbf{5.1} &11.6 & \textbf{7.8} &\textbf{18.2} & 10.0 &20.8 & 3.1 & 6.9 & \textbf{3.5} & 7.2& 5.9 &12.9 \\
   & Q2P (Ours)  & \textbf{5.1} & \textbf{12.1}  & 7.4& \textbf{18.2}& \textbf{10.2} &\textbf{21.4}  & \textbf{3.3}  & \textbf{7.0}  & 3.4& \textbf{7.6}& \textbf{6.0}  & \textbf{13.3} \\        
  \midrule
 \multirow{2}{*}{Average}
  & BetaE  & 8.2&17.9 & 10.1 & 22.5& 9.6&20.1  & 4.4&9.8 &  6.4 &13.5  &7.8&16.7 \\
  & Q2P (Ours)  & \textbf{10.5}  & \textbf{21.2}  & \textbf{12.6}  & \textbf{26.4}  & \textbf{10.1}  & \textbf{20.8}  & \textbf{5.6}   & \textbf{11.9}  &\textbf{8.1} & \textbf{16.1} & \textbf{9.4}  &\textbf{19.3}  \\

\bottomrule
\end{tabular}
\end{sc}
\end{center}
 \vspace{-0.3cm}
\caption{The MRR and Hit@10 results for complement queries used by \citet{ren2020beta}. Beta Embedding is the only baseline model that can answer queries with the complement operation.   
}
 \vspace{-0.1cm}
\label{tab:negation_beta}
 
\end{table*}

\section{Experiments}

The experiments in this section demonstrate the effectiveness and efficiency of Query2Particles.

\subsection{Experimental Setup}
The Query2Particles method is evaluated on three commonly used knowledge graphs, FB15K \cite{bordes2013translating}, FB15K-237 \cite{toutanova2015observed}, and NELL995 \cite{carlson2010toward} with the standard training, validation, and testing edges separations. 
For each of these graphs,  the corresponding training graph $G_{train}$, validation graph $G_{valid}$, and testing graph $G_{test}$ are created from training edges, training + validation edges, and training + validation + testing edges respectively.

There are two sets of complex logical queries sampled from these knowledge graphs, and the existing methods evaluate their performance on either of them. 
Specifically, \citet{ren2020query2box} sample nine different types of existential positive first-order (EPFO) queries. 
For these queries, five types of them (\texttt{1p}, \texttt{2p}, \texttt{3p}, \texttt{2i}, \texttt{3i}) are used for training and evaluation in a supervised setting. 
For the rest of four types of queries (\texttt{2u}, \texttt{up}, \texttt{ip}, \texttt{pi}), they do not appear in the training set and are directly evaluated in a zero-shot way. 
In another work, \citet{ren2020beta} refine these queries by raising the difficulties of the existing nine types of queries. 
They also include five types of complement queries (\texttt{2in}, \texttt{3in}, \texttt{inp}, \texttt{pni}, \texttt{pin}) for general first-order logic (FOL) queries.
These complement queries are also trained and evaluated in the supervised setting, but their training samples are fewer than other types. 
More details about the knowledge graphs and sampled queries are shown in the appendix. 
To demonstrate the performance of Query2Particles, it is evaluated on both sets of queries.
Note that, the query-answer pairs used for training are only from the training graph $G_{train}$. 
For validation and testing, only the hard answers from validation graph  $G_{valid}$ and testing graph $G_{test}$ are evaluated.

\subsection{Baselines}

The Query2Particles model is compared with the following baselines in the following sections.

\textbf{Graph Query Embedding (GQE)}  answers conjunctive logic queries by encoding the logical queries into vectors \cite{hamilton2018embedding}. 

\textbf{Query2Box (Q2B)} answers existential positive first-order logic queries by encoding them into boxes in the embedding space \cite{ren2020query2box}.

\textbf{Beta Embedding (BetaE)}  answers first-order logic queries by modeling them as Beta Distributions  \cite{ren2020beta}. 
This is the current state-of-the-art model on first-order logic queries.

The reported mean reciprocal rank (MRR) scores of these baselines are used by the BetaE paper \citep{ren2020beta}, and Query2Particles (Q2P) is evaluated following with the same metrics under the filtered setting, 
in which the rankings of answers are computed excluding all other correct answers.
Meanwhile, the Q2P method is also compared with other methods on EPFO queries with the queries used by \citet{ren2020query2box}.

\textbf{Continuous Query Decomposition (CQD)}  decomposes the complex queries to multiple atomic queries that can be solved by link predictors \cite{arakelyan2020complex} . 

\textbf{Embedding Query Language (EmQL)} improves the faithfulness in the reasoning process by encoding EPFO queries into centroid-sketch representations \cite{sun2020faithful}. 

The reported Hit@3 results of these two baselines are used by \citet{arakelyan2020complex, sun2020faithful}. Our model is evaluated on FB15K, FB15K-237, and NELL in the same setting.

\begin{table*}[t]
\begin{center}
\small
\begin{sc}
\begin{tabular}{p{1.8cm}||p{0.6cm}p{0.6cm}p{0.6cm}|p{0.6cm}p{0.6cm}|p{0.6cm}p{0.6cm}|p{0.6cm}p{0.6cm}||c||c||c}
\toprule
& \multicolumn{10}{c||}{FB15K} &\multicolumn{1}{c||}{FB15K-237}&\multicolumn{1}{c}{NELL}\\
\midrule
Model& 1p&2p&3p&2i&3i&Ip&Pi&2u&Up&Avg&Avg&Avg\\
\midrule
 EmQL &42.4 &50.2 &45.9 &63.7 &70.0& \underline{60.7} &61.4& 9.0 &\underline{42.6}& 49.5& \underline{35.8} & \underline{46.8} \\
  ~~$-$ sketch & 50.6 & 46.7 & 41.6 & 61.8 & 67.3&54.2&53.5&21.6&40.0&48.6&35.5&\underline{46.8} \\
  CQD-Beam &\textbf{91.8}&\textbf{77.9}&\underline{57.7}&\underline{79.6}&\underline{83.7}&37.5&\underline{65.8}&\textbf{83.9}&34.5&\underline{68.0} & 29.0 & 37.6 \\
  CQD-CO & \textbf{91.8}&45.4&19.1&\underline{79.6}&\underline{83.7}&33.6&51.3&81.6&31.9&57.6  & 27.2 & 36.8\\
  Q2P (Ours) &\underline{90.2}&\underline{74.6}&\textbf{73.4} &\textbf{86.0} &\textbf{89.6} &\textbf{63.7} &\textbf{77.6} &\underline{83.4}&\textbf{52.7} &\textbf{76.8} &\textbf{43.0} & \textbf{52.2} \\
\bottomrule
\end{tabular}
\end{sc}
\end{center}
 \vspace{-0.3cm}
\caption{The Hit@3 results for existential positive first-order queries originally used by \citet{ren2020query2box} and the comparisons are made against the state-of-the-art baselines including EmQL and CQD methods. The best results are marked in \textbf{bold} and the second-best ones are marked with \underline{underlines}. The full results are in the appendix. 
}
 \vspace{-0.3cm}
\label{tab:epfo_box}
 
\end{table*}


\subsection{Implementation Details}
The Query2Particles model is trained on the queries in an end-to-end manner. 
To fairly compare with previous methods, we set the same size of embedding vectors as four hundred. 
We use the validation queries to tune hyperparameters for our model by using grid search. 
In the grid search, we consider the batch size from $\{1024, 2048, 4096, 8192\}$, dropout rate from $\{0.1, 0.2, 0.3\}$, learning rate from $\{10^{-4}, 3*10^{-4}, 10^{-3}\}$, and label smoothing from $\{0.3, 0.5, 0.7\}$.
The final hyperparameters are shown in the supplementary materials.
Our experiments are conducted on Titan Xp with PyTorch 1.8, and they are repeated three times. 

\subsection{Comparison with Baselines}
First, we compare Query2Particles (Q2P)  with GQE, Q2B, and BetaE on the first-order logic queries used by \citet{ren2020beta}.
The results on all fourteen types of queries are reported in Table \ref{tab:epfo_beta} and Table \ref{tab:negation_beta}. 
To fairly compare with the baseline methods, we keep the same number of parameters used in each type of query embedding.

As shown in Tables \ref{tab:epfo_beta} and \ref{tab:negation_beta},  the Q2P model can achieve more accurate results than GQE, Q2B, and BetaE on all types of queries except \texttt{2u}. 
As we keep the number of query embedding parameters the same, it indicates that the structure of particle embeddings is more suitable for encoding complex queries than boxes or Beta distributions. 

Though it is slightly less accurate on the \texttt{2u} queries, 
Q2P is more efficient in encoding the queries that include union operations. 
This is because Q2P is the first embedding method that directly models the union operation. 
To avoid direct modeling of the union operation, all previous embedding methods pre-process the queries by converting them to DNF forms.
However, the DNF forms can be exponentially larger than the original queries, and the conversion also takes exponential time.
Meanwhile, BetaE proposes to use De Morgan's law to replace one union operation with one intersection and three complements, but this substitution still largely increases the query complexity.
Instead, Q2P directly models the union operation without any pre-processing or additional parameterization, while achieving the state-of-the-art performance on \texttt{up}, which is more complicated and involving the Union operation.

\begin{figure}[t]
\begin{center}
\includegraphics[clip, trim=0.9cm 0.3cm 0.3cm 0.4cm,width=\linewidth]{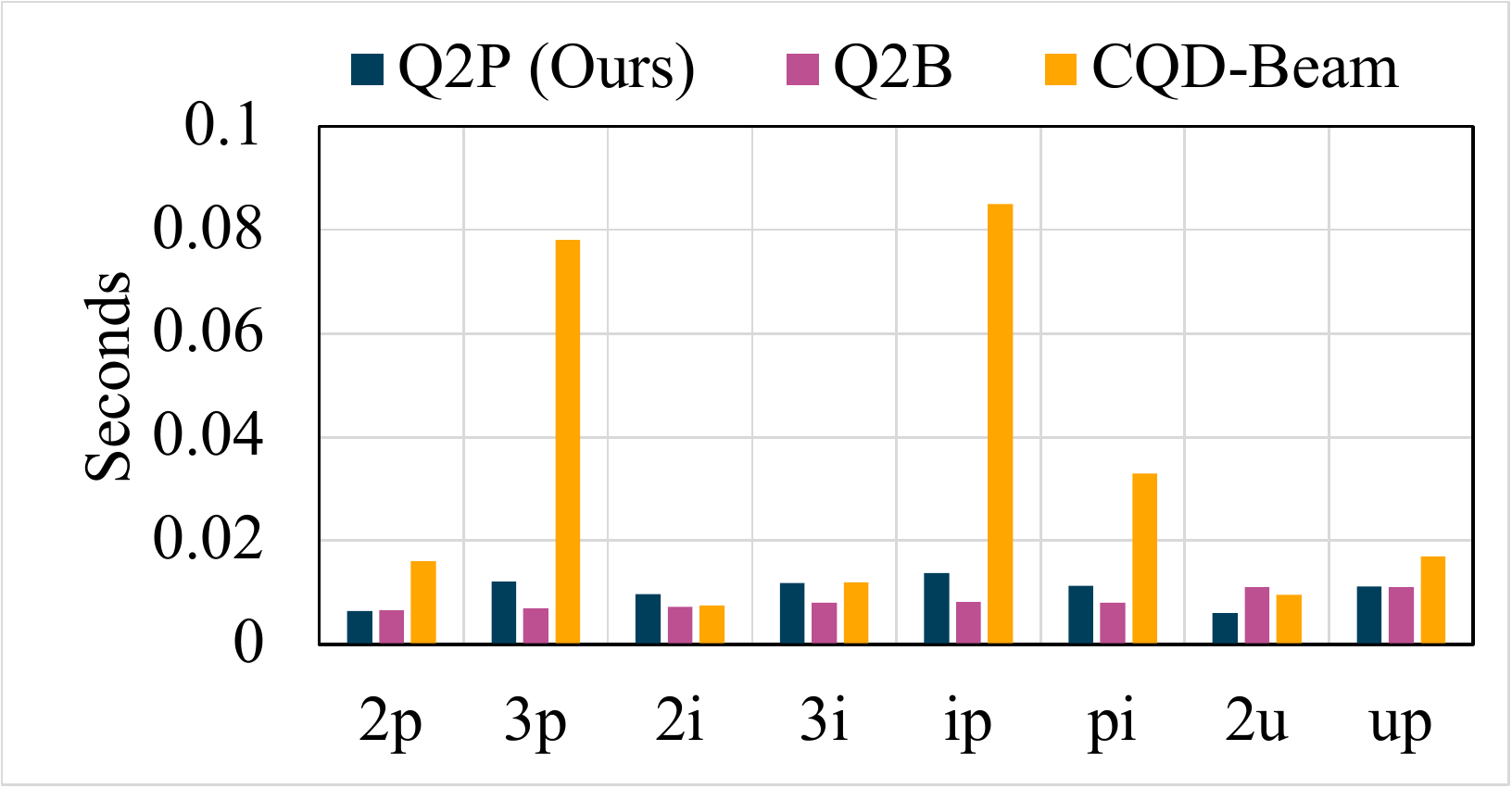}
\end{center}
\vspace{-0.3cm}
\caption{
The inference time of different types of queries on FB15k, which has the largest number of edges among three graphs. 
}
\vspace{-0.4cm}
\label{fig:timing}
\end{figure}

We also compare our model with EmQL and CQD methods on the queries used by \citet{ren2020query2box}. On average, our model has better Hit@3 scores on all datasets\footnote{In this paper, we only focus on the inductive setting, so we skip the comparison with EmQL under the entailment setting, in which the test graph is used for both training and testing.}.
Compared to the CQD method, the Q2P method is better at answering multi-hop queries.  encodes the complex queries into centroid-sketch representations, 
which cannot compactly encode sufficiently diverse answers. 
The Q2P method specifically addresses the diversity of answers, so it has higher empirical performance.
CQD performs better on shorter queries like \texttt{1p}, \texttt{2p}, and \texttt{2u}, because it can use the state-of-the-art link predictors.
Also, as shown in Figure \ref{fig:timing}, the Q2P method demonstrates a faster inference speed than the CQD method on multi-hop queries, because  CQD  uses inference time optimization, which is either a continuous optimization or a beam search.
The inference time optimization simplifies the learning of CQD but also slows down the inference efficiency on large graphs.

\subsection{The Improvement of \textsc{Q2P-Kp}}
Experiments show that the performance of the diversified queries can be largely improved by using more particles.
To demonstrate the effects, we conduct additional evaluations on the most diversified 10\% queries for each query type,
as shown in the \textsc{Divr} columns in Table \ref{tab:num_particles}. 
In doing so, we use the number of answers to measure the diversity of each query. 
In the same table, we also present the original results in the \textsc{Full} columns as a comparison.

We can observe that there is a significant performance gap between the \textsc{Full} and \textsc{Divr} results, which demonstrates that the diversified queries are harder to answer.
Meanwhile, it is also observed that
comparing to \textsc{Q2P-1p}, \textsc{Q2P-Kp (K$>$1)} significantly improves
the MRR of \textsc{Divr} queries by 7.8 points.
From this perspective, the improvement of \textsc{Q2P-Kp (K$>$1)} over \textsc{Q2P-1p} is significant on those challenging queries.

\subsection{Further Ablation Study for Q2P-1P} 
To better explain the superior performance of Q2P-1P over the baseline models, we conduct further ablations studies in Table \ref{tab:ablation_study}.

First, we remove all the self-attention layers  \texttt{Attn}.
Then the performance of intersection operations largely decreased.
This can be explained that the self-attention structure is important for aggregating the information from multiple sub-queries. 

Then, we remove all the neural network structures, including all \texttt{MLP} and \texttt{Attn} from all operations, and replace them with the operations defined in the GQE model \cite{hamilton2018embedding}. 
Then the performance of Q2P is also reduced.
This indicates that the neural structures in the particle operations are also important to the overall improvement.
Thus, we infer that the baseline model underfit the complex queries in the training set, 
and the performance can be improved by introducing more parameters and non-linearity. 
This conclusion is also aligned with \citet{sun2020faithful}, in which they found the baselines cannot faithfully answer the queries that are observed in the training time. 

However, solely using more complex structures cannot address the problem raised from the diversity of the answers. 
As shown in Table \ref{tab:num_particles}, on the top of \textsc{Q2P-1p},  \textsc{Q2P-Kp (K$>$1)} can still largely improve the performance on the diversified queries.

\begin{table}[t]
\begin{center}

\sc

\tiny
\begin{tabular}{p{0.8cm}||p{ 0.18cm}p{ 0.27cm}|p{ 0.18cm}p{ 0.27cm}|p{ 0.18cm}p{ 0.27cm}|p{ 0.18cm}p{ 0.27cm}||p{ 0.18cm}p{ 0.27cm}}

\toprule
\multirow{2}{*}{Models}  & \multicolumn{2}{c}{1p}  & \multicolumn{2}{c}{2i}  & \multicolumn{2}{c}{2u} & \multicolumn{2}{c}{2in} & \multicolumn{2}{c}{Average}   \\
 & Full &Divr & Full & Divr &Full & Divr & Full & Divr & Full & Divr\\
 \midrule
Q2P-1p  & 81.8 & 44.8 & 63.4&28.8 & 33.4& 11.3 & 18.9& 15.0 &49.4 & 25.0 \\
Q2P-2p  & 82.6& 49.4 & \textbf{65.1 }& 35.5 & 32.1 & 13.3 & \textbf{21.9}& 20.7 & 50.4 & 29.7 \\
Q2P-3p  &  \textbf{82.9} &\textbf{53.0}& 64.4 &  \textbf{37.7} &\textbf{33.6} & \textbf{18.6} & 21.8 & \textbf{21.6} & \textbf{50.6} & \textbf{32.8} \\
\bottomrule
\end{tabular}
\end{center}

\caption{The \textsc{Full} columns demonstrate the averaged MRR results, and the \textsc{Divr} columns demonstrate the averaged MRR on the top ten-percent diversified queries. 
}
\label{tab:num_particles}

\end{table}

\begin{table}[t]
\begin{center}
\sc
\tiny
\begin{tabular}{p{2.5cm}||p{ 0.4cm}|p{ 0.4cm}|p{ 0.4cm}|p{ 0.4cm}|p{ 0.4cm}}
\toprule
Models & 1p  & 2p  & 2i  & 2u  & 2in   \\
\midrule

Q2P-Kp  & \textbf{83.4} & \textbf{31.5} &\textbf{66.0} & \textbf{38.9} & \textbf{22.3} \\
Q2P-1p  & 81.8 & 30.7 & 63.4 & 33.4 & 18.9 \\
~~$-$ Self Attention  & 78.5 &28.5 & 30.9 & 30.3& 15.2 \\
~~$-$ All NNs $+$ GQE Ops & 56.7 & 16.1& 39.2& 20.1& $-$\\
\midrule
Q2P  & 68.0& 21.0& 55.1& 35.1& $-$ \\
GQE  & 54.6& 15.3 & 39.7& 22.1& $-$  \\

\bottomrule
\end{tabular}
\end{center}
\vspace{-0.3cm}
\caption{The ablation study on the Q2P neural networks. 
The \textsc{Q2P-KP} result shows the highest result when \textsc{K} is ranged from 2 to 6.
}
\label{tab:ablation_study}
\vspace{-0.4cm}
\end{table}

\section{Conclusion}

In this paper, we proposed Query2Particles, a query embedding method for answering complex logical knowledge graph queries over incomplete knowledge graphs. 
The Query2Particle method supports a full set of FOL operations.
Specifically, the Q2P method is the first query embedding method that can directly model the union operation without any preprocessing.
Experimental results show that the Q2P method achieves state-of-the-art performances on answering FOL queries on three different knowledge graphs while using comparable inference time as the previous methods. 

\section{Ethical Impacts}
This paper introduces a knowledge graph reasoning method, and the experiments are on several publicly available benchmark datasets. As a result, there is no data privacy concern. 
Meanwhile, this paper does not involve human annotations, and there is no related ethical concerns.

\section{Acknowledgements}
The authors of this paper were supported by the NSFC Fund (U20B2053) from the NSFC of China, the RIF (R6020-19 and R6021-20) and the GRF (16211520) from RGC of Hong Kong, the MHKJFS (MHP/001/19) from ITC of Hong Kong and the National Key R\&D Program of China (2019YFE0198200) with special thanks to Hong Kong Mediation and Arbitration Centre (HKMAAC) and California University, School of Business Law \& Technology (CUSBLT), and the Jiangsu Province Science and Technology Collaboration Fund (BZ2021065).

\bibliography{custom}
\bibliographystyle{acl_natbib}

\appendix

\begin{table*}[!htbp]
\sc
\begin{center}
\begin{small}
\begin{tabular}{l||cc||ccc||c}
\toprule
\textbf{Dataset} & \textbf{Relations} & \textbf{Entities} & \textbf{Training Edges} & \textbf{Validation Edges} & \textbf{Testing Edges} & \textbf{All Edges}\\
\midrule
FB15k & 1,345 & 14,951 & 483,142 & 50,000 & 59,071 & 592,213\\
FB15k-237 & 237  & 14,505& 272,115 & 17,526 & 20,438 & 310,079\\
NELL995 & 200  & 63,361& 114,213 & 14,324 & 14,267 & 142,804\\
\bottomrule
\end{tabular}
\caption{The basic information about the three knowledge graph used for the experiments, and their standard training, validation, and testing edges separation according to \cite{ren2020beta}.}
\label{tab:KG_details}

\end{small}
\end{center}
\vskip -0.1in
\end{table*}

\begin{table*}[!htbp]
\sc
\begin{center}
\begin{small}
\begin{tabular}{l||c|c|c|c}
\toprule
\textbf{Dataset} & \textbf{Batch Size} & \textbf{Dropout Ratio} & \textbf{Label Smoothing} & \textbf{Learning Rate}\\
\midrule
FB15k & 8,192 & 0.1 & 0.5 & 0.001\\
FB15k-237 & 8,192  & 0.1& 0.5 & 0.001\\
NELL995 & 4,096  & 0.3 & 0.7 & 0.0003\\
\bottomrule
\end{tabular}
\caption{The best hyperparameters used by the Query2Particles model for the experiments on the queries originally used by \cite{ren2020beta}.}
\label{tab:hyper_parameters}

\end{small}
\end{center}
\vskip -0.1in
\end{table*}

\begin{table*}[!htbp]
\begin{center}
\begin{sc}
\small
\begin{tabular}{l||cc||cc||cc}

\toprule
\cite{ren2020query2box} & \multicolumn{2}{c|}{\textbf{Training}} & \multicolumn{2}{c|}{\textbf{Validation}} & \multicolumn{2}{c}{\textbf{Test}} \\
\midrule
\textbf{Dataset}     & 1p     & Others   & 1p    & Others  & 1p     & Others  \\ \midrule
FB15k     & 273,710  & 273,710  & 59,097  & 8,000   & 67,016   & 8,000   \\
FB15k-237 & 149,689  & 149,689  & 20,101  & 5,000   & 22,812   & 5,000   \\
NELL995     & 107,982  & 107,982  & 16,927  & 4,000   & 17,034   & 4,000   \\ 
\midrule
\cite{ren2020beta} & \multicolumn{2}{c|}{\textbf{Training}} & \multicolumn{2}{c|}{\textbf{Validation}} & \multicolumn{2}{c}{\textbf{Test}} \\
\midrule
\textbf{Dataset}     & 1p/2p/3p/2i/3i     & 2in/3in/Inp/Pin/Pni   & 1p    & Others  & 1p   & Others  \\ \midrule
FB15k     & 273,710  & 273,71  & 59,097  & 8,000   & 67,016   & 8,000   \\
FB15k-237 & 149,689  & 149,68  & 20,101  & 5,000   & 22,812   & 5,000   \\
NELL995     & 107,982  & 107,98  & 16,927  & 4,000   & 17,034   & 4,000   \\ 
\bottomrule

\end{tabular}

\end{sc}
\end{center}

\caption{The detailed information for the queries used for training, validating, and testing all query embedding methods. 
The upper parts disclose the statistics of the queries taken from the \cite{ren2020query2box} paper, while the lower part describes the queries taken from \cite{ren2020beta}. 
The major differences are that the queries in \cite{ren2020beta} is harder than \cite{ren2020query2box}, and include five additional types of queries with the complement operation.
}
\label{tab:queries_details}
\end{table*}




\begin{table*}[ht!]
\begin{center}
\small
\begin{sc}
\begin{tabular}{p{1.8cm}|p{2.2cm}||p{0.7cm}p{0.7cm}p{0.7cm}|p{0.7cm}p{0.7cm}|p{0.7cm}p{0.7cm}|p{0.7cm}p{0.7cm}||p{0.7cm}}
\toprule
Dataset&Model& 1p&2p&3p&2i&3i&Pi&Ip&2u&Up&Avg\\
\midrule
\multirow{4}{*}{FB15k}
  & BetaE & 65.1 & {25.7} & {24.7} & {55.8} &  {66.5} &  {43.9} & {28.1} & \textbf{{40.1}}  & 25.4 & {41.6} \\
  & Q2B & {68.0} & 21.0 & 14.2 & 55.1 & {66.5} & 39.4 & 26.1 & 35.1  & 16.7 &  38.0 \\
  & GQE  & 54.6 & 15.3 & 10.8 & 39.7 & 51.4 & 27.6 & 19.1 & 22.1  & 11.6 &  28.0 \\
  & Q2P (Ours) & \textbf{82.6} & \textbf{30.8} & \textbf{25.5} & \textbf{65.1}& \textbf{74.7} & \textbf{49.5}& \textbf{34.9} &32.1& \textbf{26.2} & \textbf{46.8} \\
\midrule
\multirow{4}{*}{FB15k-237} 
  & BetaE & 39.0 & {10.9} & {10.0} & 28.8 & {42.5} & {22.4} & {12.6} & \textbf{{12.4}}  &  \textbf{{9.9}} & {20.9} \\
  & Q2B & \textbf{{40.6}} & 9.4 & 6.8 & {29.5} & 42.3 & 21.2 & {12.6} & 11.3  & 7.6  & 20.1 \\
  & GQE & 35.0 & 7.2 & 5.3 & 23.3 & 34.6 & 16.5 & 10.7 & 8.2  & 5.7  & 16.3 \\
  & Q2P (Ours) &39.1 &\textbf{11.4} &\textbf{10.1}&\textbf{32.3}&\textbf{47.7}&\textbf{24.0}&\textbf{14.3}&8.7&9.1&\textbf{21.9} \\
\midrule
 \multirow{4}{*}{NELL}
  & BetaE & {53.0} & 13.0 & {11.4} & \textbf{{37.6}} & {47.5} & \textbf{{24.1}} & 14.3 & \textbf{{12.2}}   & 8.6 & {24.6} \\
  & Q2B & 42.2 & {14.0} & 11.2 & 33.3 & 44.5 & 22.4 & {16.8} & 11.3  & {10.3}  & 22.9 \\
  & GQE & 32.8 & 11.9 & 9.6 & 27.5 & 35.2 & 18.4 & 14.4 & 8.5  & 8.8 & 18.6 \\
  & Q2P (Ours) &\textbf{56.5}&\textbf{15.2}&\textbf{12.5}&35.8&\textbf{48.7}&22.6&\textbf{16.1}&11.1&\textbf{10.4}&\textbf{25.5} \\
  \midrule
 \multirow{4}{*}{Average}
  & BetaE & 52.4 & 16.5 & 15.4 & 40.7 & 52.2& 30.1 & 14.3 &\textbf{21.6}  & 14.5 & 29.1 \\
  & Q2B &50.3&14.8&10.7&39.3&51.1&27.7&18.5&19.2&11.5&27.0\\
  & GQE & 40.8 &11.5&8.6& 30.2& 40.4 & 20.8& 14.7 & 12.9& 8.7 & 21.0 \\
  & Q2P (Ours) & \textbf{59.4} & \textbf{19.1} & \textbf{16.0} & \textbf{44.7} & \textbf{57.0 }& \textbf{32.0}& \textbf{21.8} & 17.3& \textbf{15.2}& \textbf{31.3 }\\
\bottomrule
\end{tabular}
\end{sc}
\end{center}
\caption{The MRR result for existential positive first order queries comparing to the BetaE, Q2B, and GQE methods. The results are reported from the queries used by \citet{ren2020beta}.
}
\label{tab:epfo_beta_full}
 
\end{table*}

\begin{table*}[ht!]
\begin{center}
\small
\begin{sc}
\begin{tabular}{p{1.8cm}|p{2.2cm}||p{0.7cm}p{0.7cm}p{0.7cm}|p{0.7cm}p{0.7cm}|p{0.7cm}p{0.7cm}|p{0.7cm}p{0.7cm}||p{0.7cm}}
\toprule
Dataset&Model& 1p&2p&3p&2i&3i&Ip&Pi&2u&Up&Avg\\
\midrule
\multirow{5}{*}{FB15k}
& EmQL &42.4 &50.2 &45.9 &63.7 &70.0& \underline{60.7} &61.4& 9.0 &\underline{42.6}& 49.5 \\
&  ~~$-$ sketch & 50.6 & 46.7 & 41.6 & 61.8 & 67.3&54.2&53.5&21.6&40.0&48.6 \\
 & CQD-Beam &\textbf{91.8}&\textbf{77.9}&\underline{57.7}&\underline{79.6}&\underline{83.7}&37.5&\underline{65.8}&\textbf{83.9}&34.5&\underline{68.0} \\
 & CQD-CO & \textbf{91.8}&45.4&19.1&\underline{79.6}&\underline{83.7}&33.6&51.3&81.6&31.9&57.6 \\
 & Q2P (Ours) &\underline{90.2}&\underline{74.6}&\textbf{73.4}&\textbf{86.0}&\textbf{89.6}&\textbf{63.7}&\textbf{77.6}&\underline{83.4}&\textbf{52.7}&\textbf{76.8 }\\
\midrule
\multirow{5}{*}{FB15k-237} 
 & EmQL &37.7&\underline{34.9}&\underline{34.3}&\underline{44.3}&\underline{49.4}&\textbf{40.8}&\underline{42.3}&8.7&28.2&35.8 \\
&  ~~$-$ sketch & 43.1 & 34.6& 33.7 & 41.0 & 45.5 & \underline{36.7} & 37.2 & 15.3 & \textbf{32.5} & \underline{35.5} \\
 & CQD-Beam &\textbf{51.2}&28.8&22.1&35.2&45.7&12.9&24.9&\underline{28.4}&12.1&29.0 \\
 & CQD-CO &\textbf{51.2}&21.3&13.1&35.2&45.7&14.6&22.2&28.1&13.2&27.2 \\
 & Q2P (Ours) &\underline{49.0}&\textbf{44.2}&\textbf{44.6}&\textbf{50.1}&\textbf{57.5}&34.1&\textbf{44.2}&\textbf{32.9}&\underline{30.6}&\textbf{43.0}\\
\midrule
 \multirow{5}{*}{NELL}
 & EmQL &41.5&\underline{40.5}&\underline{38.6}&\textbf{62.9}&\textbf{74.5}&\textbf{49.8}&\textbf{64.8}&12.6&35.8&\underline{46.8} \\
&  ~~$-$ sketch &48.3 & 39.5 & 35.2 & \underline{57.2} & \underline{69.0} & \underline{48.0} & \underline{59.9} & 25.9 &\textbf{38.2}& \underline{46.8} \\
 & CQD-Beam &\underline{66.7}&35.0&28.8&41.0&52.9&17.1&27.7&\textbf{53.1}&15.6&37.6 \\
 & CQD-CO &\underline{66.7}&26.5&22.0&41.0&52.9&19.6&30.2&\textbf{53.1}&19.4&36.8 \\ & Q2P (Ours)&\textbf{67.0}&\textbf{53.0}&\textbf{52.6}&52.9&\underline{69.0}&38.0&47.0&\underline{52.9}&\underline{37.0}&\textbf{52.2}\\
 \midrule
 \multirow{5}{*}{Average}
 & EmQL&40.5&41.9&\underline{39.6}&\underline{57.0}&\underline{64.6}&\textbf{50.4}&\underline{56.2}&10.1&35.5&44.0 \\
&  ~~$-$ sketch &47.3 &40.3&36.8&53.3&60.6&\underline{46.3}&50.1&20.9&\underline{36.9}&43.6\\
  & CQD-Beam &\textbf{69.9}&\underline{47.2}&36.2&51.9&60.8&22.5&39.5&\underline{55.1}&20.7&\underline{44.9} \\ 
  & CQD-CO &\textbf{69.9}&31.1&18.1&51.9&60.8&22.6&34.6&54.3&21.5&40.5 \\
 & Q2P (Ours)&\underline{68.7}&\textbf{57.3}&\textbf{56.9}&\textbf{63.0}&\textbf{72.0}&45.3&\textbf{56.3}&\textbf{56.4}&\textbf{40.1}&\textbf{57.3}\\

\bottomrule
\end{tabular}
\end{sc}
\end{center}
\caption{The Hit@3 results for existential positive first order queries comparing to the EmQL and CQD method over the queries used by \citet{ren2020query2box}. 
}
\label{tab:epfo_box_full}
 
\end{table*}

\end{document}